\documentclass{article}

\usepackage{arxiv}

\usepackage[utf8]{inputenc} 
\usepackage[T1]{fontenc}    
\usepackage{hyperref}       
\usepackage{url}            
\usepackage{booktabs}       
\usepackage{amsfonts}       
\usepackage{nicefrac}       
\usepackage{microtype}      
\usepackage{lipsum}		
\usepackage{graphicx}
\usepackage{natbib}
\usepackage{doi}
\usepackage{lineno,hyperref}
\usepackage{times}
\usepackage{soul}
\usepackage{url}
\usepackage[small]{caption}
\usepackage{graphicx}
\usepackage{amsmath}
\usepackage{amsthm}
\usepackage{booktabs}
\usepackage{algorithm}
\usepackage{algorithmic}
\usepackage{multirow}
\usepackage{subfigure}
\usepackage{color,xcolor}
\usepackage{subfigure} 
\usepackage{wrapfig}

\title{DCF-ASN:  Coarse-to-fine Real-time Visual Tracking via Discriminative Correlation Filter and Attentional Siamese Network}

\date{31 January, 2020}	

\author{ {\hspace{1mm}Xizhe Xue}\\
	School of Computer Science\\
	Northwestern Polytechnical University\\
	Xi'an, China \\
	\texttt{xuexizhe@mail.nwpu.edu.cn} \\
	\And
	{\hspace{1mm}Ying Li}\thanks{Corresponding author.}  \\
	School of Computer Science\\
	Northwestern Polytechnical University\\
	Xi'an, China \\
	\texttt{lybyp@nwpu.edu.cn} \\
	\And
	{\hspace{1mm}Xiaoyue Yin}  \\
	School of Computer Science\\
	Northwestern Polytechnical University\\
	Xi'an, China \\
	\texttt{2015302412@mail.nwpu.edu.cn} \\
	\And
	{\hspace{1mm}Qiang Shen} \\
	Department of Computer Science\\
	Aberystwyth University\\
	Aberystwyth, SY23 3DB, UK \\
	\texttt{qqs@aber.ac.uk} \\
}

\renewcommand{\shorttitle}{DCF-ASN:  Coarse-to-fine Real-time Visual Tracking}

\hypersetup{
	pdftitle={DCF-ASN:  Coarse-to-fine Real-time Visual Tracking via Discriminative Correlation Filter and Attentional Siamese Network},
	pdfsubject={Computer vision},
	pdfauthor={Xizhe Xue},
	pdfkeywords={Visual tracking, Correlation filter, Feature fusion},
}

\begin{document}
\maketitle
\begin{abstract}
		Discriminative correlation filters (DCF) and siamese networks have achieved promising performance on visual tracking tasks thanks to their superior computational efficiency and reliable similarity metric learning, respectively. However, how to effectively take advantages of powerful deep networks, while maintaining the real-time response of DCF, remains a challenging problem. Embedding the cross-correlation operator as a separate layer into siamese networks is a popular choice to enhance the tracking accuracy. Being a key component of such a network, the correlation layer is updated online together with other parts of the network. Yet, when facing serious disturbance, fused trackers may still drift away from the target completely due to accumulated errors. To address these issues, we propose a coarse-to-fine tracking framework, which roughly infers the target state via an online-updating DCF module first and subsequently, finely locates the target through an offline-training asymmetric siamese network (ASN). Benefitting from the guidance of DCF and the learned channel weights obtained through exploiting the given ground-truth template, ASN refines feature representation and implements precise target localization. Systematic experiments on five popular tracking datasets demonstrate that the proposed DCF-ASN achieves the state-of-the-art performance while exhibiting good tracking efficiency.
\end{abstract}


\section{Introduction}

Visual tracking, which tracks an arbitrary temporally changing object based on the speciﬁed ground truth at the ﬁrst frame, plays an active role in a wide range of applications, including robotics, surveillance, human-computer interaction and motion analysis. However, fast motion, partial or full occlusion, background clutter and many other factors in the image sequences make it a challenge to perform effective and efficient tracking. How to estimate the target state in complex scenarios with balanced accuracy and speed is the core problem of this task.

\begin{figure}[!htb]
\centering
\subfigure[Sequence \textit{S1607}]{\includegraphics[width=0.48\textwidth]{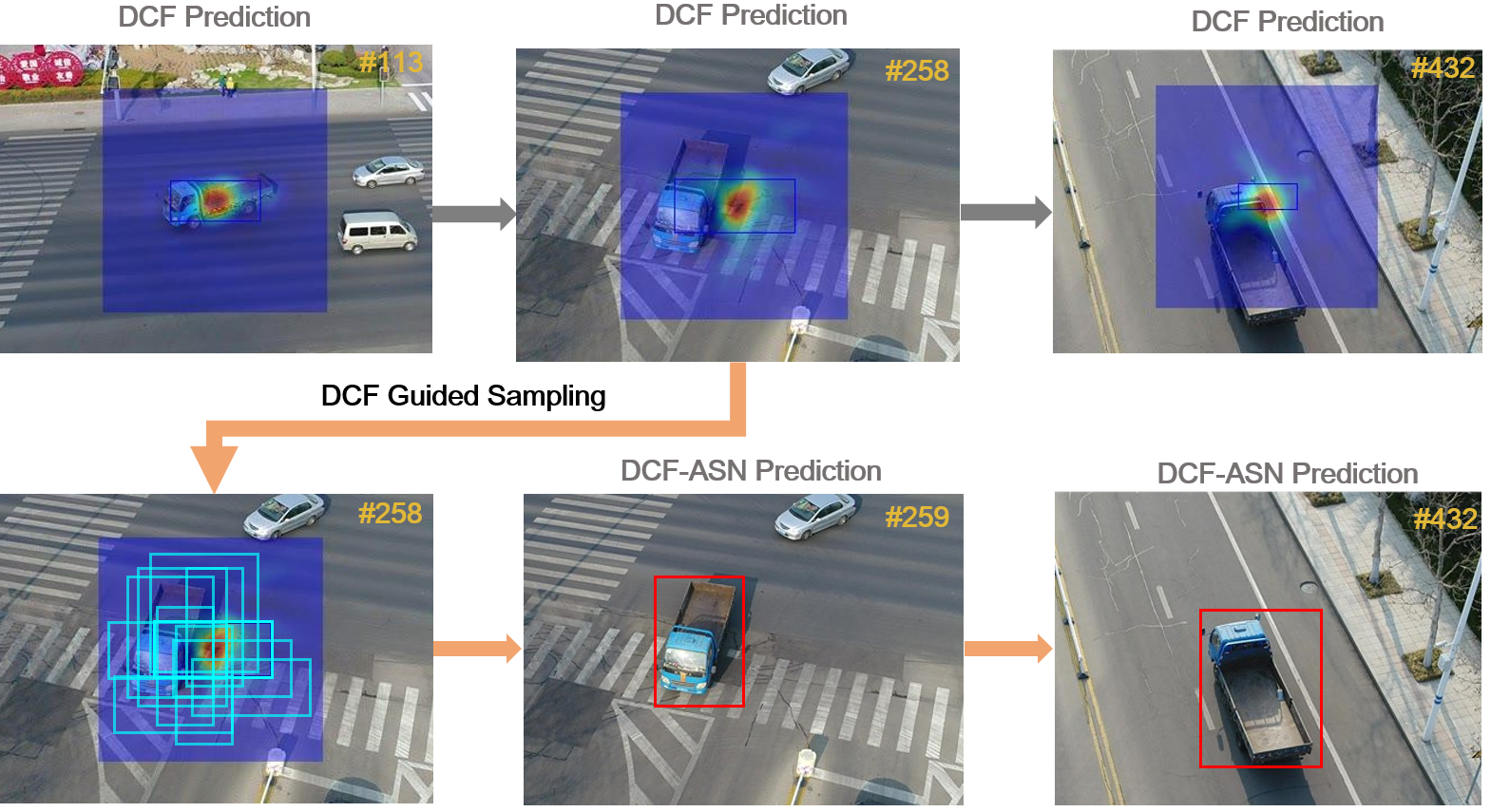}} 
\subfigure[Sequence \textit{Girl2}]{\includegraphics[width=0.48\textwidth]{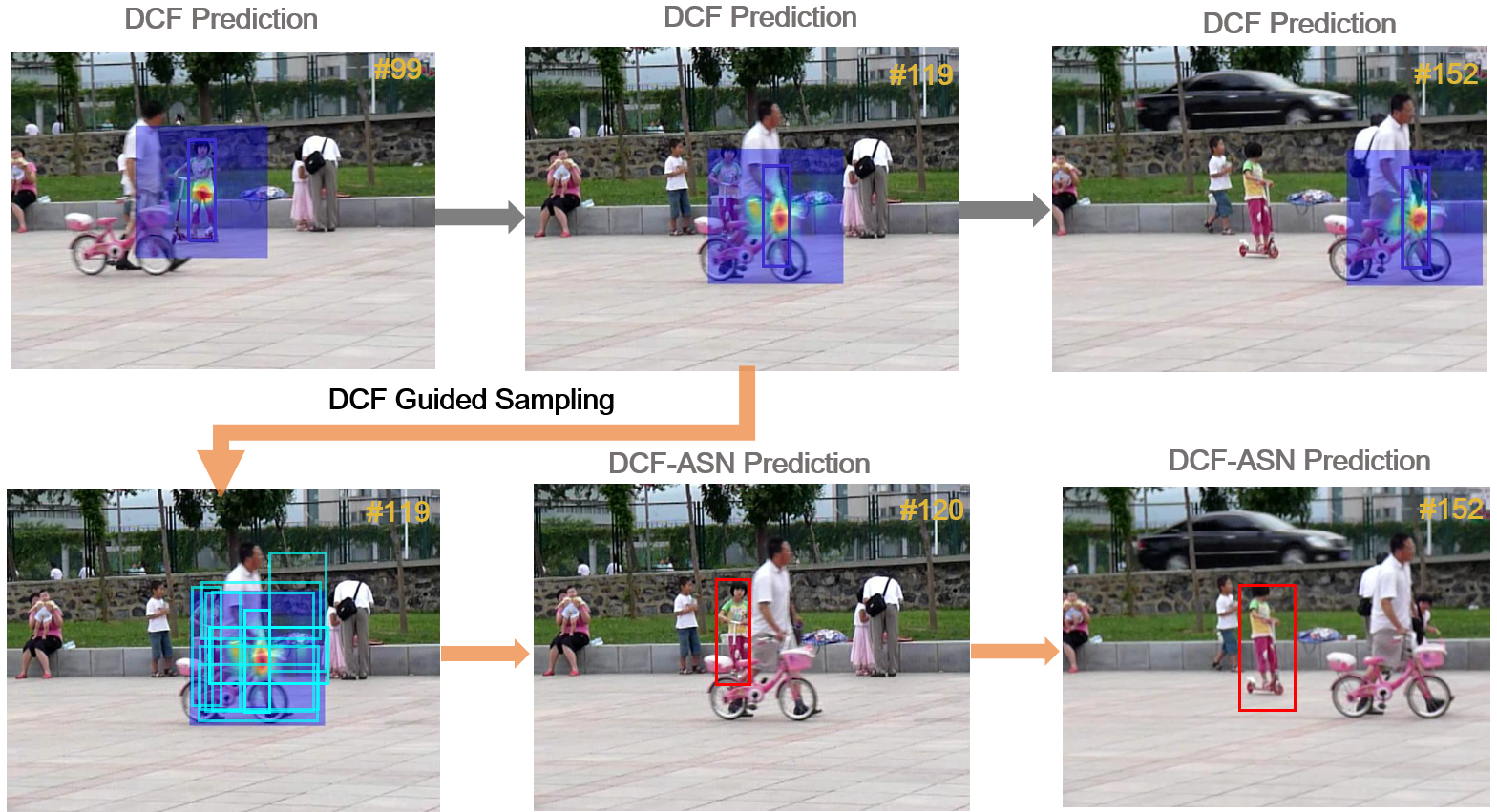}} 
\caption{Illustration of "coarse-to-fine" stratgy on sequences \textit{S1607} and  \textit{Girl2} obtained from datasets UAVDT and OTB100. After DCF guided sampling,  proposed DCF-ASN is able to relocate the  target missed by the DCF.}
\label{fig:fig1}
\vspace{0.2in}
\end{figure}

Benefiting from  cycle shift operation and fast Fourier transform (FFT), discriminative correlation filter (DCF)~\cite{huang2020transfer} based trackers are popular for their efficiencies among the existing tracking algorithms. They learn to discriminate an image patch from the surrounding patches by solving a ridge regression problem and calculating the correlation response in the frequency domain. More recently, research on visual tracking has focused on deep learning based methods~\cite{li2018deep,chen2020siamese,danelljan2020probabilistic}. Much attention has been paid into siamese networks~\cite{xiao2020memu}, which always extract deep features and then locate the target by measuring the similarity of extracted features. Although trained through the large scale datasets, the early siamese trackers have not been able to achieve state-of-the-art (SOTA) performance. The follow-on methods as reported in~\cite{valmadre2017end,danelljan2019atom,bhat2019learning} obtain certain improvements by employing online-updating rules and embedding a cross-correlation layer into siamese structure.  However, such an online-updating tracker based on end-to-end siamese network may drift away if inevitable tracking errors become serious. Recent work as per siamese-RPN~\cite{li2018high} found that an offline trained deep tracker could also achieve competitive results without online adaptation. Inspired by these observations, an efficient coarse-to-fine tracking framework is proposed in this paper, where the online updating DCF module first predicts target position coarsely and the offline trained siamese network performs fine target localization. The contribution of this work is three-fold: 

1. We present a coarse-to-fine visual tracking framework via an online updating DCF and an offline learned siamese network, in which two modules correct and promote each other (See Figure \ref{fig:fig1}). 

2. We propose a novel attentional siamese network (ASN) to learn the appropriate channel weights from a given template through off-line training, offering more powerful and robust weighted feature maps for accurate bounding box regression.

3. The proposed tracking framework achieves competitive performance on 7 popular benchmarks of OTB100, VOT2018, LaSOT, GOT10K, TrackingNet, UAV123 and UAVDT, while performing at real-time speed. Both qualitative and quantitative analyses demonstrate of robustness of the proposed tracker.

\section{Related Work}
{\noindent \bf Discriminative correlation filter based tracking:} DCF trackers perform fast tracking through circular convolution, which can be implemented efficiently through transformation to the frequency domain. The first application of correlation filters for tracking was presented in ~\cite{bolme2010visual}, which worked by manipulating the maximum cross-correlation response between the target and the candidate patches. Following this initial approach, a number of improvements over the DCF trackers have been introduced, including those based on: kernel space~\cite{henriques2014high}, multi-scale filters~\cite{danelljan2014accurate}, traditional fused features~\cite{bertinetto2016staple}, spatial-regularization operates~\cite{li2018learning}, and factorized convolution~\cite{danelljan2017eco}. In light of the effectiveness of deep converlution neural networks (CNNs), other algorithms~\cite{danelljan2017eco,wang2018multi} have been developed that experience an increase on tracking accuracy when employing a pretrained deep network to produce features ~\cite {he2016deep}. However, the full potential of deep CNNs in performing tracking tasks has not been exploited by the offline trained feature extractor. 

{\noindent \bf Siamese network based tracking:} Siamese trackers each consist of two branches, which encoder different patches into feature maps and compare their similarities in the implicitly embedded space. Inspired by correlation based methods, the pioneering work SiamFC~\cite{bertinetto2016fully} introduces a cross-correlation layer, thereby achieving beyond real-time speed. However, there remains a significant gap between SiamFC and the SOTA techniques in terms of online adaptability. As a follow-up work, CFNet~\cite{valmadre2017end} breaks through this bottleneck by constructing an end-to-end online updating siamese network. Aiming to be adaptable to the appearance variation of the target, DSiam~\cite{guo2017learning} employs fast transformation learning and multi-layer fusion, which supports adaptively integrating the network outputs. Apart from these approaches, other popular optimization methods include those that exploit triplet loss ~\cite {dong2018triplet}, introduce residual attention learning ~\cite{wang2018learning} and anchor-free mechanism~\cite{chen2020siamese}, and impose the width and depth of a network~\cite{zhang2019deeper} or semantic mask branch~\cite {wang2019fast}. Additionally, learning from object detection methods, there exist further approaches~\cite{danelljan2020probabilistic,danelljan2019atom,li2018high} that consider the tracking task as a one-shot detection problem, which perform well with the assistance of large-scale training image pairs.
\begin{figure}[!htb]
\begin{center}
	\includegraphics[width=0.95\linewidth]{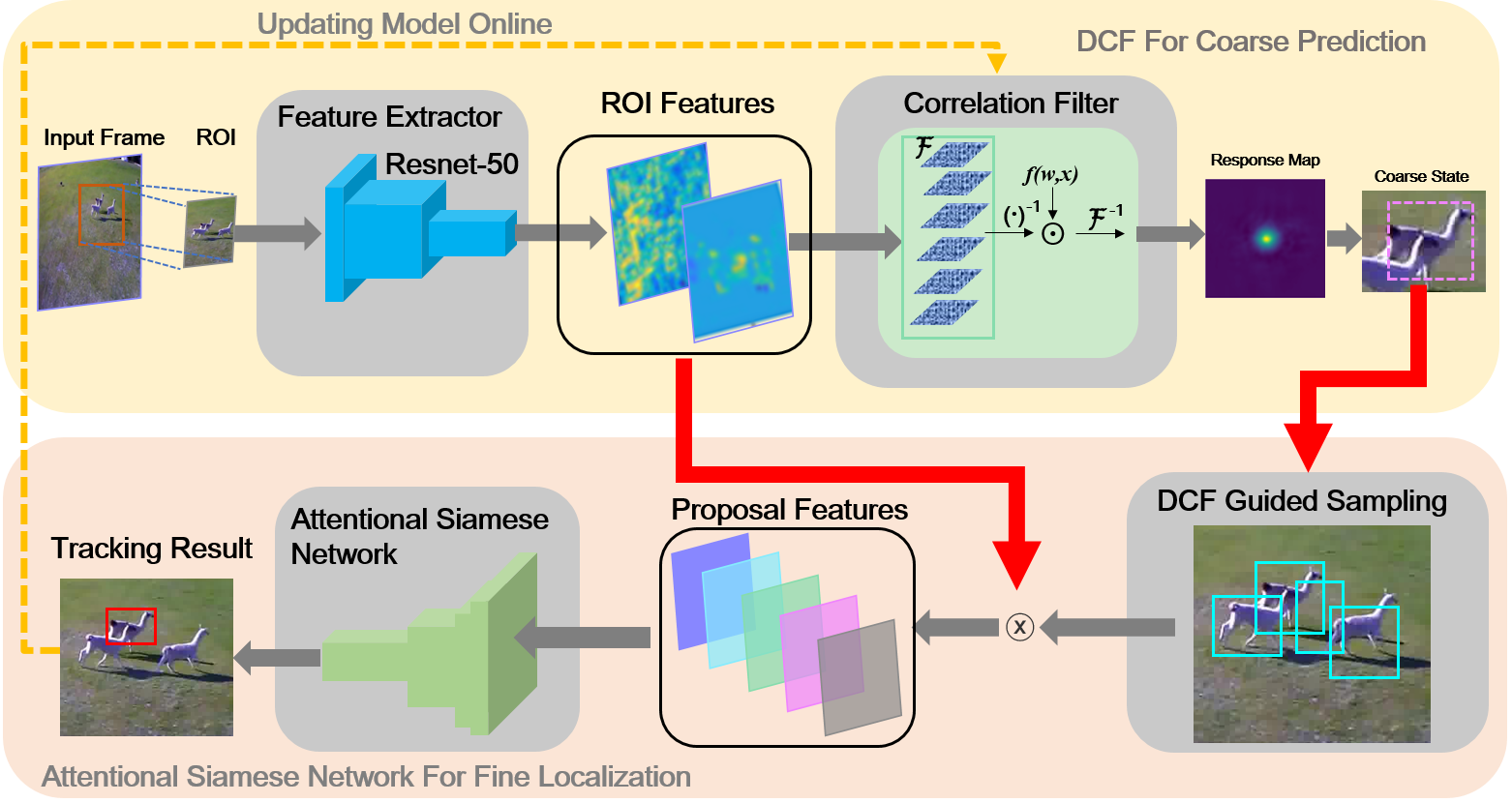}
\end{center}
\caption{Architecture of proposed coarse-to-fine visual tracking framework. Region of interest (ROI) patch from the original frame is fed into this framework directly. A standard DCF is utilized to roughly predict the target state (location and scale) in the ROI. Then, proposals of different scales are generated around the position resulting from the DCF module. Cropped from the ROI feature maps, proposal features are fed into a novel attentional siamese network in order to locate the target accurately.}
\label{fig:fig2}
\end{figure}
\section{Coarse-to-fine Tracking Framework}
In order to fully adapt to the change of target appearance while suppressing potential model drift caused by online updating, we propose a novel coarse-to-fine tracking framework. As displayed in Figure \ref{fig:fig2}, it consists of two components: a discriminative correlation filter for coarse target prediction and a class-agnostic attentional siamese network (ASN) for precise localization. The target estimation network in ATOM~\cite{danelljan2019atom} is adopted as the baseline to design the proposed ASN. The complete procedure of DCF-ASN is shown in Algorithm 1.

\begin{algorithm}[!htb]
\caption{Tracking algorithm}
\label{alg:algorithm}
\textbf{Input}: Video frames $f_1,f_2,...,f_i$ of length $L$\\
Initial target state $S_1$\\
\textbf{Output}: Target state $S_i$ in frame $f_i$
\begin{algorithmic}[1] 
	\STATE	Crop $S_1$ from $f_1$ and resize it to multiple scale patches $R_{_1}^T\left( {T = \{ 1,2,3,4,5\} } \right)$.\\
	\STATE Extract multi-scale features $x_1$ from  $R_{_1}^T$ and input it into the proposed attentional siamese network (ASN).\\
	\WHILE{every $f_i$ $(i>1)$  in $L$}
	\STATE Sample the new patch ${Z^{{f_i}}}$ based on the size of ${S_{i - 1}}$ and resize it to multiple scale patches $R_{_i}^T$.\\
	\STATE Extract multi-scale features $x_i$ from $R_{_i}^T$ and calculate the response of DCF filter with (1) and (2).\\
	\STATE 	Predict the coarse position $P_i$ (the location with highest value in response map) and the corresponding scale $C_i$ of target.\\
	\STATE	Add some Gaussian perturbation on the position $P_i$ and scale $C_i$ to generate $N$ proposals around the coarse location.\\
	\STATE Crop features of proposal regions from $x_i$ and send them into ASN.\\
	\STATE	Rank the proposals according to the score obtained from ASN.\\
	\STATE	Find the proposal with highest score and consider it as the fine target state $S_i$.\\
	\STATE Update DCF: ${w_{i + 1}} = (1 - lr) * {w_{i - 1}} + lr * {w_i}$.\\
	\ENDWHILE
	\STATE \textbf{return} $S_i$ $(i=1,2,...,L)$	
\end{algorithmic}
\end{algorithm}
\subsection{DCF based coarse target prediction}
The DCF module is utilized to roughly estimate the target state in the proposed framework, and the structure of the filter itself is very simple in an effort to reduce the additional computation. Sharing the same feature extractor with ASN, the DCF module represents the $i-th$$(i \in[1, D])$ layer feature of sample  $j-th$$(j \in[1, N])$ as $x_{ji}$ , and the online adaptation process can be viewed as that of minimizing the following loss function:
\begin{equation}
\resizebox{.8\linewidth}{!}{$
\displaystyle
l(w; x)=\sum_{j=1}^{N} \sum_{i=1}^{D} \mu_{j}\left\|f\left(w, x_{j i}\right)-y_{j i}\right\|_{}^{2}+\sum_{i=1}^{D}\left\|\lambda f^{i}\right\|_{}^{2}
$}
\end{equation}%
where $w$ denotes the parameter vector of the filter $f$,  $y_{ji}$ symbolizes the desired response,   $\lambda$ represents the regularization parameter to avoid boundary effects, and ${\mu _j}$ controls the impact of each training sample.

By applying FFT to convert the learned filter from the temporal to the frequency domain, we can rewrite (1) as:
\begin{equation}
\resizebox{0.8\linewidth}{!}{$
\displaystyle
L(w; x)=\sum_{j=1}^{N} \sum_{i=1}^{D} \mu_{j}\left\|F\left(W, X_{ji}\right)-Y_{ji}\right\|_{}^{2}+\sum_{i=1}^{D}\left\|\lambda * F^{i}\right\|_{2}^{2}
$}
\end{equation}%
where each capital letter stands for the corresponding form of the original temporal variable.

Making full use of the sparsity structure of this problem as (2), the DCF module proposed herein utilizes the Conjugate Gradient method (CG)~\cite{danelljan2017eco} to resolve it efficiently, computing the target state in the frequency domain. After transforming the optimal result back to the time domain by inverse FFT, the resulting position and scale are considered as the coarse prediction, around which proposals for fine location are sampled according to Gaussian distribution. Because regions that are closer to the coarse location always owns the higher probability of target  existence, more proposals need to be sample there. Meanwhile, the further away  should also be covered with a small number of bounding boxes.
\begin{figure}[!htb]
\begin{center}
	\includegraphics[width=1\linewidth]{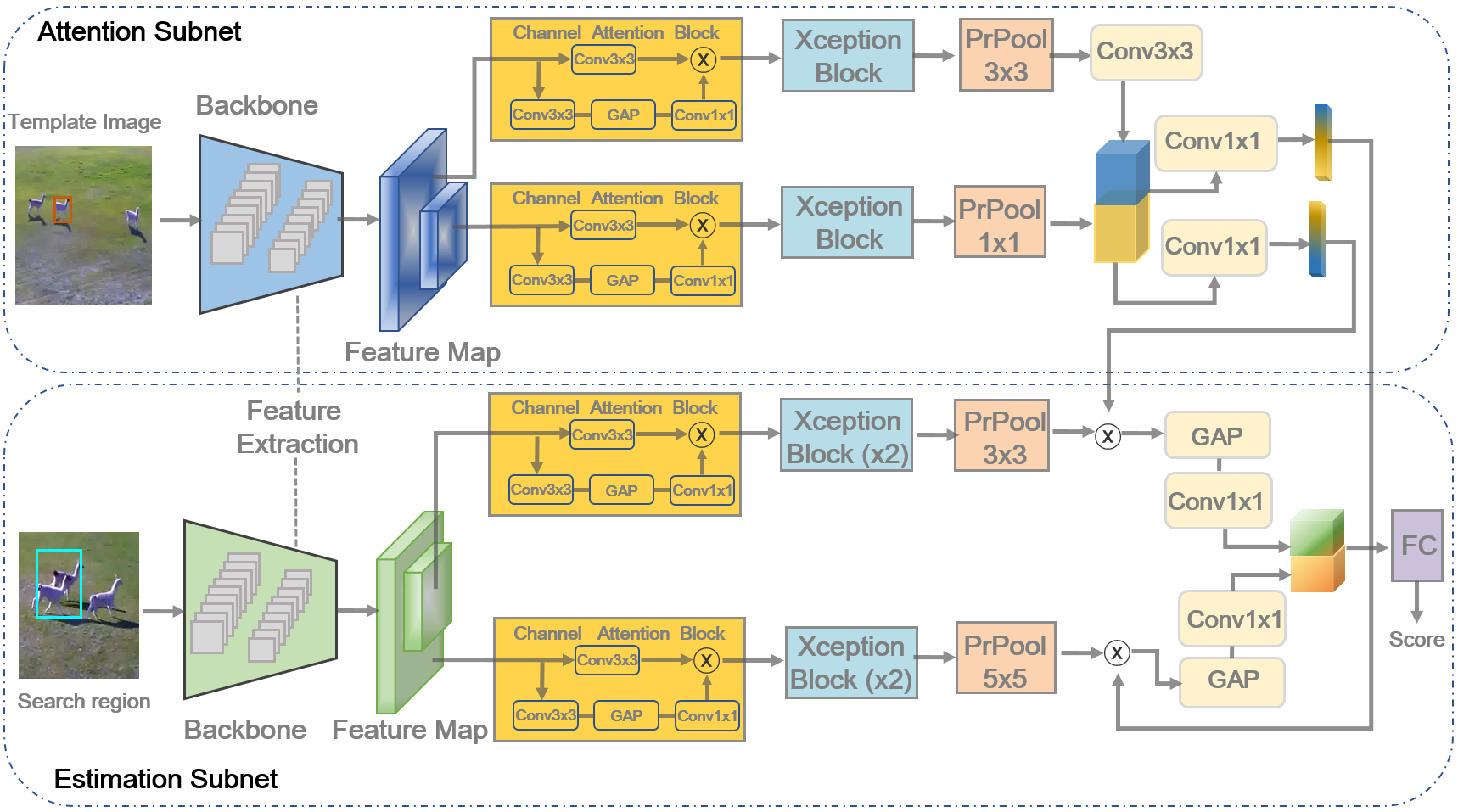}
\end{center}
\caption{Proposed ASN for fine localization. The first frame with a given ground truth in sequence is considered as the template image while the proposal boxes in the search region of the current frame are sampled under the guidance of DCF result. The backbone in each subnet is based on a pre-trained Resnet-50 (with feature maps extracted from Res-block3 and Res-block4). The Xception blocks in the proposed framework are of the same structure as that given in \protect\cite{chollet2017xception}, and GAP represents the Global Average Pooling operation. The $ \otimes $ symbol means element-wise product between the current feature maps and the corresponding channel weights produced by the attention subnet.}
\label{fig:fig3}
\end{figure}
\subsection{ASN for fine target localization}
Based on the coarse state of the target estimated by the DCF module, we intend to perform a precise prediction by capturing the latent relations between the given template image and the current target. The majority of existing methods work by extracting features from both the template image and the search region, and the place with the highest feature similarity provides the hint for the target location. However, inspired by the prior investigations of~\cite{valmadre2017end,wang2018learning}, we construct an attentional siamese network, in which an attention subnet learns the channel weights of the features concerned and an estimation subnet utilizes the refined feature maps to predict the score of a certain proposal. Finally, among various candidate proposals, the one with the largest score indicates the fine location estimated from the ASN. The components of each subnet are shown in Figure \ref{fig:fig3}.
\paragraph{\textbf{Attention Subnet}} This subnet starts with a backbone that extracts two feature maps of different depths from the template image. Each feature map is refined by a simple channel attention block followed by a classical Xception block with its dimensionality reduced to a quarter of the input. Then, the Precise ROI Pooling is utilized to compress the spatial size of the feature maps. A 1x1 pooling operates on the smaller feature map, while a 3x3 spatial pooling and a 3x3 convolution are sequentially introduced together on the lager feature map to avoid the loss of local details. After that, the two resulting feature vectors are fused and convoluted by two 1x1 convolutional filters to generate the channel weights.
\paragraph{\textbf{Estimate Subnet}} Sharing the same backbone with the attention subnet, the estimate subnet owns a channel attention block as well. What is different is that the two Xception blocks are contained within each branch of the estimate subnet and the Precise ROI Pooling operators here reduce the spatial size of two feature maps to 3x3 and 5x5, respectively. Subsequently, features in different channels are re-weighted and the resulting features are transformed to vectors through GAP and 1x1 convolution, the two reﬁned feature vectors are concatenated and fed into an FC-layer to predict the final confidence score of each proposal.
\paragraph{\textbf{Training and Loss Function}} In order to train the proposed ASN, we apply the mean-squared error (MSE) loss function $L_{ASN}$ for accurate target state prediction, as in (3).
\begin{equation}
\resizebox{0.8\linewidth}{!}{$
\displaystyle
{L_{ASN}}({P_{gt}},{P_{mn}},Score_{mn}) =  {\left\| {GIOU\left( {{P_{{\rm{gt}}}},{P_{mn}}} \right) - Scor{e_{mn}}} \right\|^2}
$}
\end{equation}%
where $Scor{e_{mn}},{P_{mn}}$  denote the $n-th$ prediction output and proposal box in frame $m$, representatively. $P_{gt}$ means the given ground truth and the GIOU~\cite{rezatofighi2019generalized} represents the generalized intersection over union between two boxes.

Particularly, the template and search image are sampled from the training sequences with a maximum gap of 50 frames. For each image pair, we generate 16 candidate proposal boxes by adding Gaussian noise to the ground truth coordinates, while ensuring a minimum GIoU of 0.1. The MSE between the predicted score and the true computed GIOU is optimized  using ADAM~\cite{danelljan2017eco}.

\section{Experiments}
In this section, the proposed DCF-ASN approach is evaluated and compared with SOTA tracking algorithms on thousands of challenging image sequences obtained from five popular visual tracking datasets: OTB100~\cite{wu2015object}, VOT2018~\cite{kristan2018sixth}, LaSOT~\cite{fan2019lasot}, GOT-10k~\cite{huang2019got} and TrackingNet~\cite{muller2018trackingnet}. All evaluation criteria are according to the original protocol deﬁned in the five benchmarks respectively.

\subsection{Implementation Details}
The proposed DCF-ASN tracker is implemented in Python under the Pytorch framework. The proposed ASN is trained on the training subset of the corresponding datasets (LaSOT, GOT-10K, TrackingNet) with 128 frames per batch. Durning the training process, the training parameters in ASN are set to be the same as those in ATOM. All experiments of each tracker are carried out on a workstation with an Intel Xeon E5-2699 processor (2.30GHz) and NVIDIA 1080ti GPU. 

\subsection{State-of-the-art Comparison}
For comparative studies on the five tracking benchmarks, we evaluate the proposed algorithm with SOTA trackers, including BACF~\cite{kiani2017learning}, STRCF~\cite{li2018learning}, MCCT~\cite{wang2018multi}, SiamFC~\cite{bertinetto2016fully}, ECO~\cite{danelljan2017eco}, CFNet~\cite{valmadre2017end}, SiamFC-tri~\cite{dong2018triplet}, VITAL~\cite{song2018vital}, SiamRPN++~\cite{li2019siamrpn++}, SiamMask~\cite{wang2019fast}, ATOM~\cite{danelljan2019atom} and DiMP~\cite{bhat2019learning}.

\begin{table}[htbp]
\footnotesize
\centering
\resizebox{0.95\linewidth}{!}{
	\begin{tabular}{p{5.em}|c|c|c|c|c|c}
		\toprule
		\multirow{2}[4]{*}{Method} & \multirow{2}[4]{*}{Backbone} & \multicolumn{2}{p{4.11em}|}{OTB100} & \multicolumn{2}{p{4.em}|}{VOT2018} & \multicolumn{1}{c}{\multirow{2}[4]{*}{FPS}} \\
		\cmidrule{3-6}    \multicolumn{1}{c|}{} & \multicolumn{1}{c|}{} & \multicolumn{1}{p{2.em}|}{AUC} & \multicolumn{1}{p{2.em}|}{Pr} & \multicolumn{1}{p{2.em}|}{EAO} & \multicolumn{1}{p{2.em}|}{Acc} &  \\
		\midrule
		\midrule
		BACF  & -     & 0.597 & 0.810  &  -  &  -  & 10 \\
		STRCF & -     & 0.524 & 0.722 & -   &  -  & 20 \\
		MCCT  & VGGNet & 0.673 & 0.898 & 0.393 & 0.580 & 2 \\
		SiamFC & AlexNet & 0.560  & 0.747 & 0.188 & 0.503 & \textcolor[rgb]{ 1,  0,  0}{122} \\
		ECO   & VGGNet & 0.659 & 0.892 & 0.280 & 0.484 & 28 \\
		SiamFC-tri & AlexNet & 0.576 & 0.765 & - & - & \textcolor[rgb]{ .357,  .608,  .835}{67} \\
		VITAL & -     & 0.678 & \textcolor[rgb]{ 1,  0,  0}{0.910} & - & - & 3 \\
		SiamRPN++ & ResNet50 & \textcolor[rgb]{ 1,  0,  0}{0.692} & \textcolor[rgb]{ .357,  .608,  .835}{0.907} & 0.415  & 0.600 & 15 \\
		SiamMask & ResNet50 & 0.643 & 0.832 & 0.380  &\textcolor[rgb]{ .357,  .608,  .835}{ 0.610} & 24 \\
		ATOM  & ResNet18 & 0.659 & 0.869 & 0.401 & 0.590 & 30 \\
		DiMP  & ResNet50 & {0.683} & 0.892 & \textcolor[rgb]{ .357,  .608,  .835}{0.440} & {0.597} & 40 \\
		\textbf{DCF-ASN} & ResNet50 &\textcolor[rgb]{ .357,  .608,  .835}{0.684}  & 0.906 & \textcolor[rgb]{ 1,  0,  0}{0.450} & \textcolor[rgb]{ 1,  0,  0}{0.612} & 38 \\
		\bottomrule
	\end{tabular}%
	\label{tab:tab1}%
}
\caption{Comparisons with state-of-the-art methods on the OTB100 and VOT2018 datasets. AUC: area under curve; Pr: precisoin; EAO: expected average overlap; Acc: accuracy; FPS: frame per second. The first and second score are marked with red and blue respectively.}
\label{tab:tab1}
\end{table}%

\paragraph{\textbf{OTB100 and VOT2018}} The OTB100 dataset consists of 100 challenging videos fully annotated with 11 different attributes. Based on the overlap precision (Pr) and area-under-the-curve (AUC) metrics, we validate the proposed DCF-ASN and other SOTA methods. As shown in Table \ref{tab:tab1}, The proposed algorithm, integrating an efficient correlation filter with an attentional siamese model, achieves competitive results on both AUC and Pr. It is obvious that the proposed DCF-ASN (which incorporates the merits of DCF module and efficient attentional siamese network ) runs at about 38 FPS on a single GPU, outdistancing the real-time requirement for pratical applications. 

VOT2018 is a widely-used supervised evaluation metric for visual object tracking, containing 60 sequences with various challenging factors. It is annotated with the rotated bounding boxes, and a re-initialized methodology is applied for evaluation. In particular, the VOT2018 benchmark measures the trackers in terms of accuracy (Acc), robustness (R), and expected average overlap (EAO). In this subsection, we compare the proposed method with the SOTA trackers reported in the VOT2018 Challenge. Table 1 shows that our tracker achieves the best performance in terms of EAO while maintaining a very competitive accuracy and speed. Although SiamMask utilizes much larger training data (including VOS dataset), our tracker can still outperform it.
\begin{figure*}[htbp]
\centering
\includegraphics[width=0.48\linewidth]{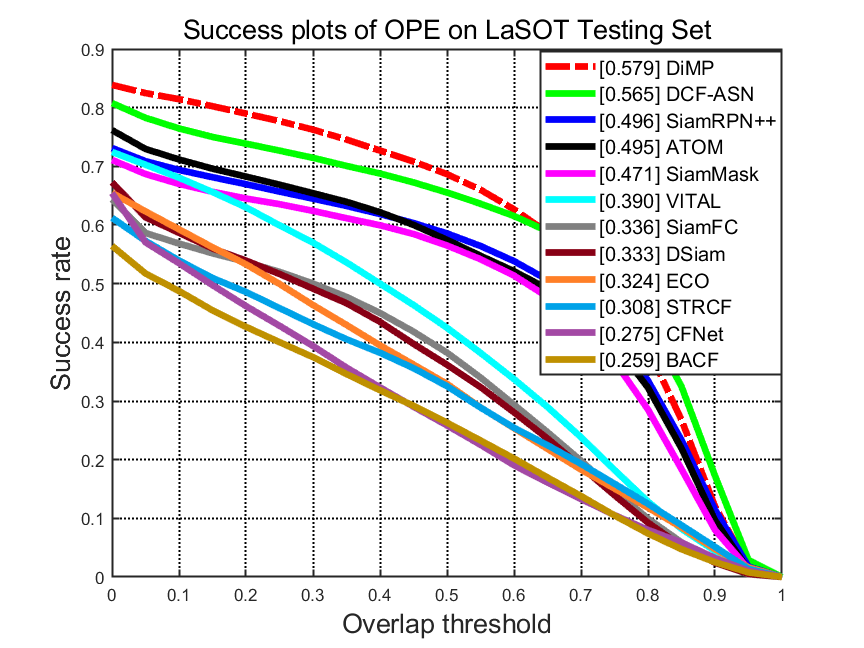}
\includegraphics[width=0.48\linewidth]{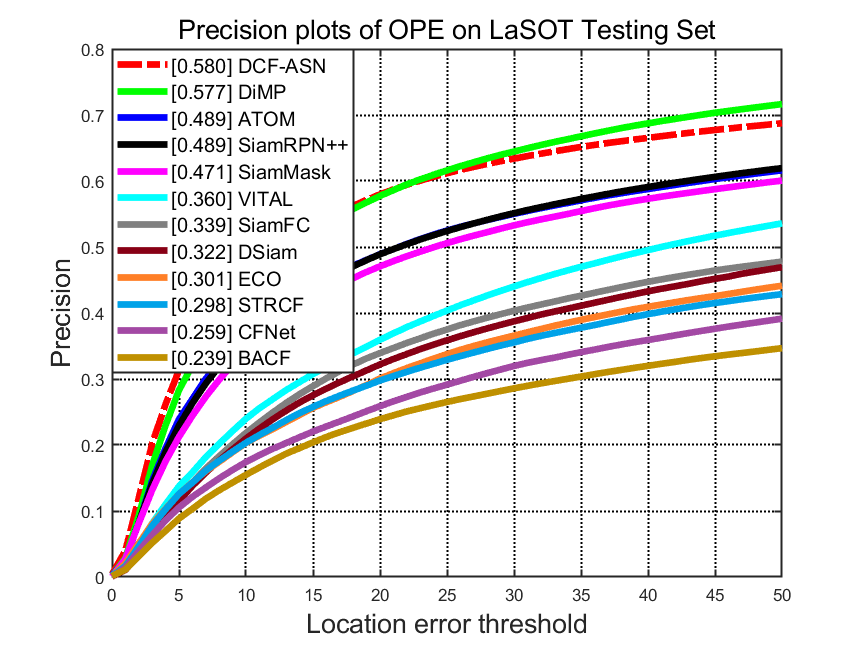}
\caption{State-of-the-art comparison on the large-scale dataset LaSOT in terms of success plot and precision plot.}
\label{fig:fig4}
\end{figure*}
\paragraph{\textbf{LaSOT}}The LaSOT dataset is a  large-scale tracking dataset, consisting of 1400 sequences with average length of 2512 frames. It covers 70 object categories, in which each categorie contains 20 different sequences. 

To provide a comprehensive evaluation, we have experimented on the testing subset with 280 videos, showing AUC in the success plot and Pr in the precision plot in Figure \ref{fig:fig4}. In the success plot, the proposed tracker achieves the second best with a score of 0.565, only 1.4\% separated from the best algorithm DiMP (0.579), while significantly outperforming the third method SiamRPN++ (0.496) with siamese structure by approximately 7\%. Furthermore, DCF-ASN gains the highest score of 0.580 in the precision plot, having an advantage of 0.3\% over the second best tracker DiMP (0.577) and a distinct advantage of 9.1\% over the third method ATOM (0.489).

\paragraph{\textbf{TrackingNet}}As the largest wild object tracking dataset, TrackingNet assembles a total of over 30k video. All the 14,431,266 frames extracted from the 140 hours of visual content are annotated with a single upright bounding box. Coping with a large variety of frame rates, resolutions, context and object classes, 30,132  videos are seleted as training dates and 511 videos  with a distribution similar to the training set are utilized to test trackers. Algorithms evaluated on the TrackingNet are ranked according to AUC, precision (Pr) and normalized precision (NPr).
\begin{table*}[htbp]
\resizebox{1\linewidth}{!}{
	\centering
	\begin{tabular}{ccccccccc}
		\toprule
		\multicolumn{1}{c}{} & \multicolumn{1}{c}{ECO} & \multicolumn{1}{c}{SiamFC} & \multicolumn{1}{c}{CFNet} & \multicolumn{1}{c}{MDNet} & \multicolumn{1}{c}{ATOM} & \multicolumn{1}{c}{SiamRPN++} & \multicolumn{1}{c}{DiMP} & \multicolumn{1}{c}{\textbf{DCF-ASN}} \\[0.5pt]
		\midrule
		Pr    & 0.492 & 0.533 & 0.533 & 0.565 & 0.648 & \textcolor[rgb]{ 1,  0,  0}{0.694} & 0.687 &\textcolor[rgb]{ .357,  .608,  .835}{0.689} \\
		NPr   & 0.618 & 0.666 & 0.654 & 0.705 & 0.771 & {0.800} &{0.801} & \textcolor[rgb]{ 1,  0,  0}{0.803} \\
		AUC     & 0.554 & 0.571 & 0.578 & 0.606 & 0.703 & {0.733} & \textcolor[rgb]{ 1,  0,  0}{0.740} & \textcolor[rgb]{ .357,  .608,  .835}{0.734} \\
		\bottomrule
	\end{tabular}%
	\label{tab:tab2}%
}	
\caption{ State-of-the-art comparison on the TrackingNet dataset in terms of Pr, NPr and AUC.}
\end{table*}%

As reported in Table 2, the proposed DCF-ASN achieves favorable performance compared to SOTA trackers. Although
DCF-ASN witnesses a mere difference of 0.5\% in Pr from the winner achieved by SiamRPN++, our method can run almost as three times faster as it.

Compared with the DiMP~\cite{bhat2019learning}  that is also built on the basis of the ATOM tracker, the results of our proposed method are still competitive. Notably, DiMP changes nothing on the target estimation network of ATOM, but increases the model size to 364MB owing to the improvement of the target classification module. However, for our DCF-ASN, due to the introduction of correlation filter for rough estimation, its overall parameter number is only about 211MB, which is approximately half of that of DiMP.
\begin{table*}[htbp]
\resizebox{1\linewidth}{!}{
	\centering
	\begin{tabular}{ccccccccc}
		\toprule
		\multicolumn{1}{c}{} & \multicolumn{1}{c}{MDNet} & \multicolumn{1}{c}{BACF} & \multicolumn{1}{c}{ECO} & \multicolumn{1}{c}{CFNet} & \multicolumn{1}{c}{SiamFC} & \multicolumn{1}{c}{ATOM} & \multicolumn{1}{c}{DiMP} & \multicolumn{1}{c}{\textbf{DCF-ASN}} \\[0.5pt]
		\midrule
		$SR_{0.5}$ & 0.303 & 0.262 & 0.309 & 0.265 & 0.353 & 0.701 &\textcolor[rgb]{ .357,  .608,  .835}{0.717} & \textcolor[rgb]{ 1,  0,  0}{0.720} \\
		$SR_{0.75}$ & 0.099 & 0.101 & 0.111 & 0.087 & 0.098 & 0.479 &\textcolor[rgb]{ .357,  .608,  .835} {0.492} & \textcolor[rgb]{ 1,  0,  0}{0.496} \\
		$AO$    & 0.299 & 0.260  & 0.316 & 0.293 & 0.348 & 0.602 & \textcolor[rgb]{ .357,  .608,  .835}{0.611} & \textcolor[rgb]{ 1,  0,  0}{0.612} \\
		\bottomrule
	\end{tabular}%
}
\caption{State-of-the-art comparison on the GOT-10k dataset in terms of $SR_{0.5}$, $SR_{0.75}$ and $AO$.}
\label{tab:tab3}%
\end{table*}%
\paragraph{\textbf{GOT-10k}} GOT-10k is a high-diversity dataset including 10k video sequences, where targets annotated frame-by-frame with bounding boxes. Our tracker are evaluated on the test subset, which contains 84 different object classes and 32 motion patterns. Trackers evaluated on test set will be reported at three metrics, the average overlap ($AO$) and success rates based on two overlap thresholds 0.5 ($SR_{0.5}$) and 0.75 ($SR_{0.75}$).

Table \ref{tab:tab3} illustrates that  DCF-ASN achieves the top-ranked results in both AO (average overlap) and SR (success rate), including $SR_{0.5}$ and $SR_{0.75}$. Compared against the existing leading methods ATOM and DiMP for example, DCF-ASN makes an improvement of 1.7\% and 0.4\% respectively, in terms of $SR_{0.75}$, while also improving in $AO$ and $SR_{0.5}$.

\subsection{Ablation Study}
\begin{figure}[htbp]
\includegraphics[width=0.49\linewidth]{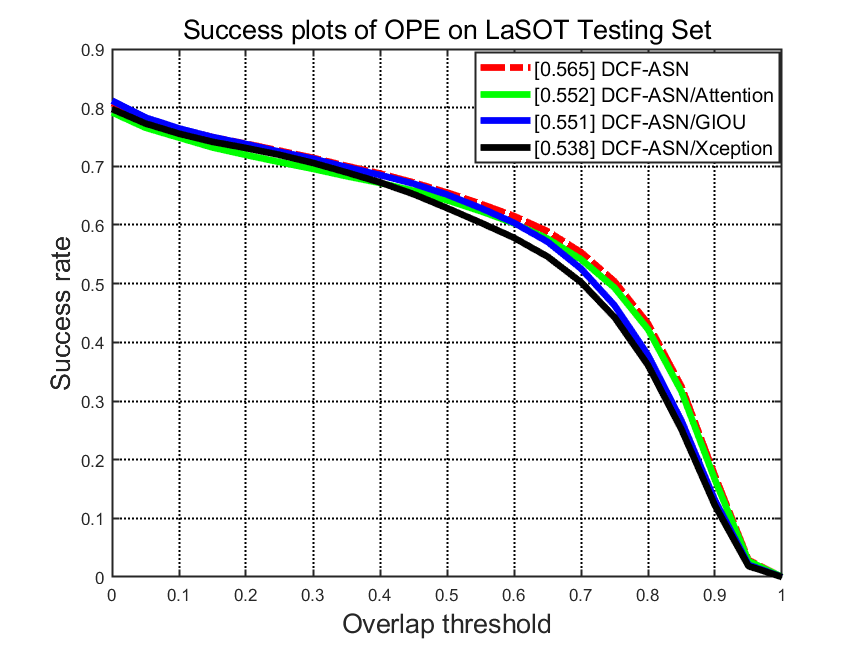}
\includegraphics[width=0.49\linewidth]{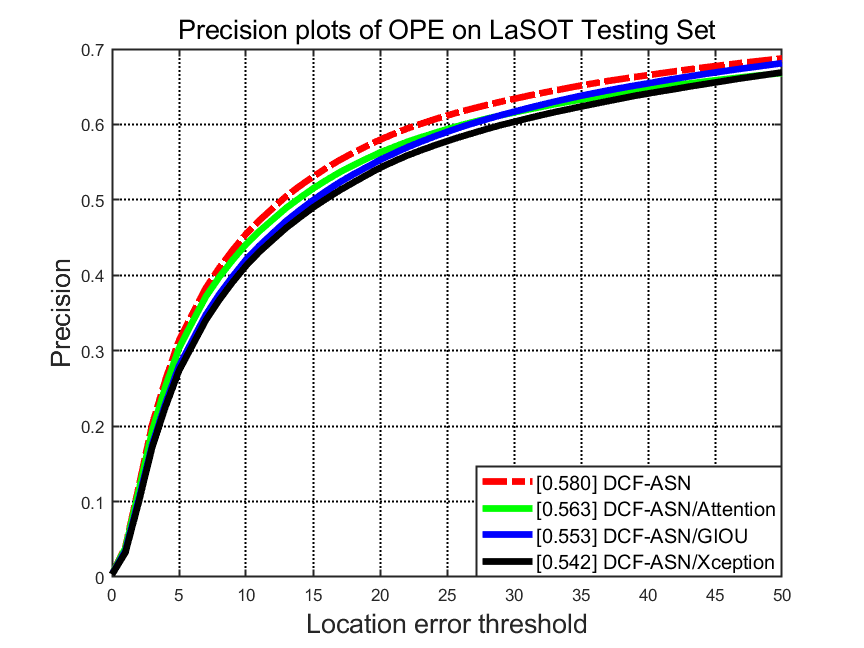}
\caption{Ablation results on LaSOT dataset in terms of AUC and Pr. Counterparts without channel attention block, GIOU, Xception block are DCF-ASN/Attention, DCF-ASN/GIOU and DCF-ASN/Xception, respectively}
\label{fig:fig5}
\end{figure}
In this part of the experimental study, we perform an extensive ablation analysis to demonstrate the impact of each components in the proposed method, further illustrating its superiority. 

As shown in Figure  \ref{fig:fig5}, the performance of DCF-ASN declines in varying degrees when different functional modules are removed. It indicates that all unique designs play an essential role in ASN. In particular, the introduction of the Xception block increases the use efficiency of the model parameters with a multiple branch structure, thereby increasing the accuracy. For the design of the attention module, we have also tried to apply Dual Attention~\cite{fu2019dual} and Criss-Cross Attention~\cite{huang2019ccnet} to capture the long range dependencies between pixels. However, the experimental results are even not so good as those attainable by the simple channel attention structure. Perhaps under the current framework, the global contextual information does not contribute much to the tracking task. In addition, experimental results also prove that GIOU are more powerful than the traditional IOU when calculating loss within the proposed framework, and this may inspire further investigation in the future. 
\begin{figure}[htbp]
\includegraphics[width=0.49\linewidth]{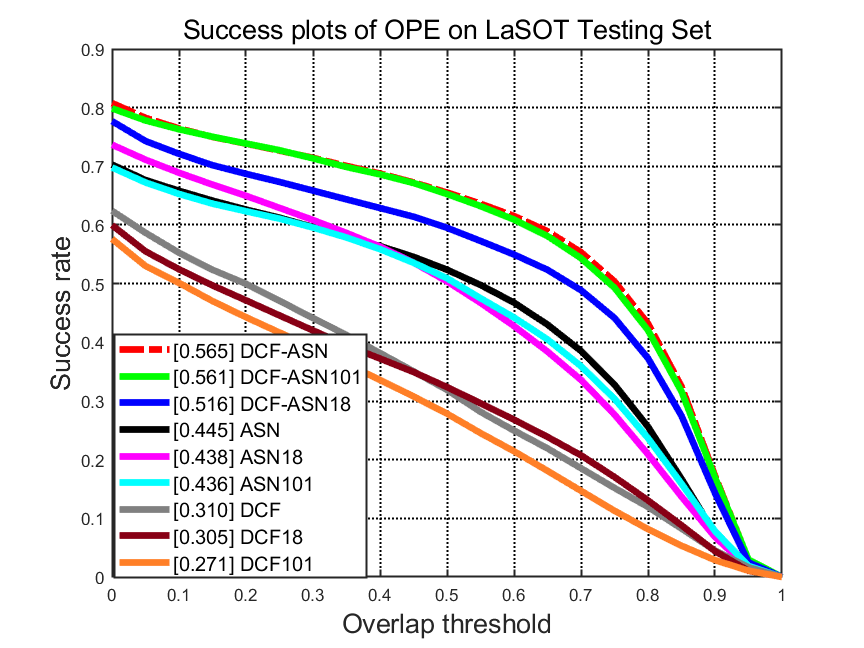}
\includegraphics[width=0.49\linewidth]{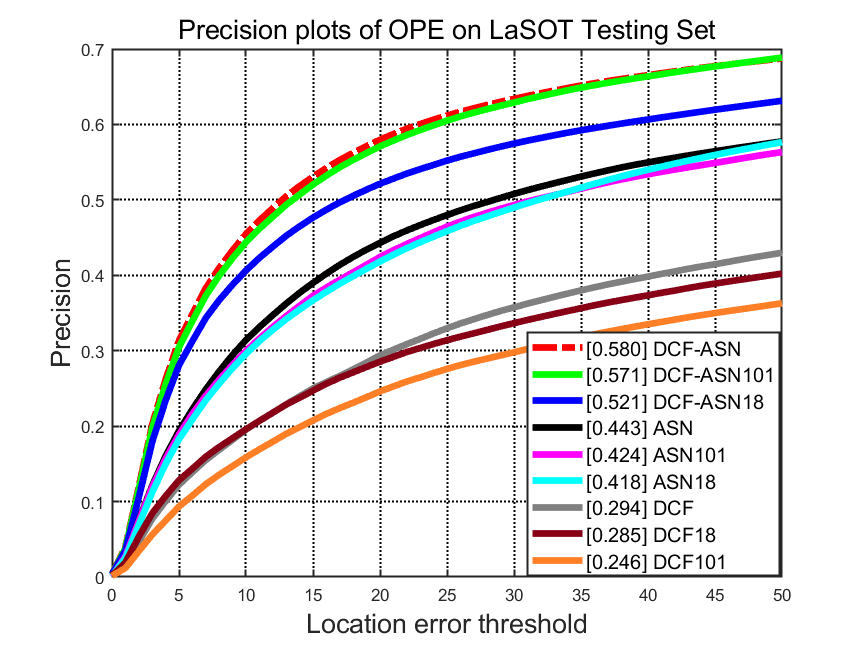}
\caption{Ablation results on LaSOT dataset in terms of AUC and Pr. DCF-ASN, DCF, ASN counterparts using backbone Resnet-18, Resnet-50 and Resnet-101 are respectively DCF-ASN(DCF, ASN)\_18, DCF-ASN(DCF, ASN)\_50 and DCF-ASN(DCF, ASN)\_101.}
\label{fig:fig6}
\end{figure}

Note that we separate DCF and ASN from the proposed DCF-ASN and test them on the LaSOT dataset. The results are shown in Figure  \ref{fig:fig6}. Thanks to the learning process over large amounts of data, the AUC and Pr of ASN are higher than those of DCF.  Nonetheless, the performance of DCF and ASN alone are much worse than that of DCF-ASN. Replacing the backbone with deeper CNN (Resnet-101) does not help to enhance the performance of the proposed method, whilst introducing more parameters.
\begin{figure*}[htbp]
\centering
\includegraphics[width=0.44\linewidth]{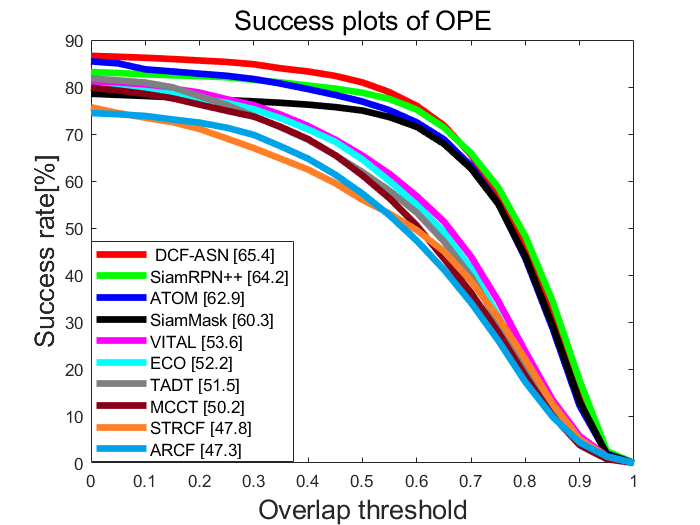}
\includegraphics[width=0.44\linewidth]{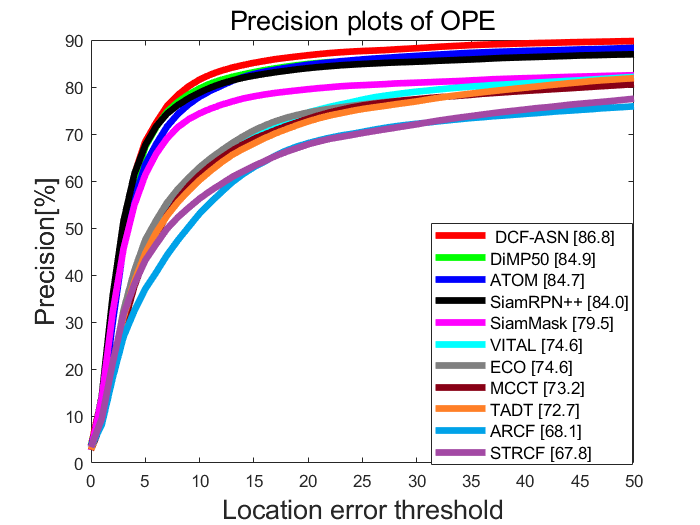}
\caption{Success and Precision plots of the proposed DCF-ASN and SOTA methods on the
	UAV123 dataset, with AUC and Precision(Pr) explicitly marked in plots.}
\label{fig:fig7}
\end{figure*}
\begin{figure*}[htbp]
\centering
\includegraphics[width=0.44\linewidth]{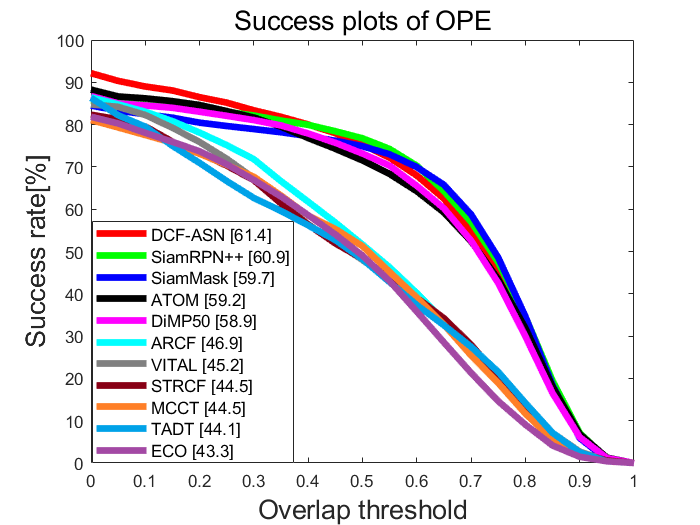}
\includegraphics[width=0.44\linewidth]{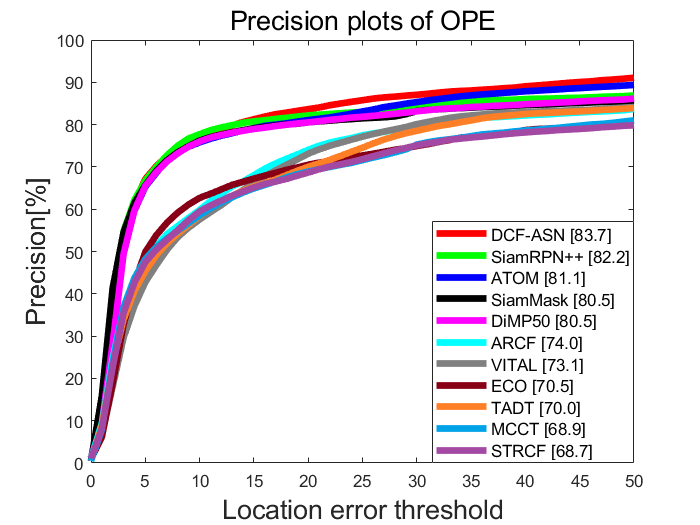}
\caption{Success and Precision plots of the proposed DCF-ASN and SOTA methods on the
	UAVDT dataset, with AUC and Precision(Pr) explicitly marked in plots.}
\label{fig:fig8}
\end{figure*}
\subsection{Evaluation under Aerial Scenario}

To further validate the generalization ability of the proposed tracker, we evaluated it in challenging aerial scenes, and compared it with representative tracking approaches on the typical datasets UAV123~\cite{mueller2016benchmark} and UAVDT~\cite{du2018unmanned}. In addition to the general SOTA algorithms, ARCF~\cite{dai2019visual}, a tracker designed specifically for aerial 
scenarios, has also been added in this comparative study.
\paragraph{\textbf{Evaluation on UAV123 Dataset}}
UAV123 is a popular aerial video-based tracking dataset that consists of 123 video sequences, of which 115 are produced by a UAV platform and 8 are created using UAV simulation software. UAV123 contains kinds of scenarios, such as fields, streets, cities, suburbs, oceans and so on. The dataset embodies a total of 12 attributes: Aspect Ratio Change, Background Clutter, Camera Motion, Fast Motion, Full Occlusion, Illumination Variation, Low Resolution, Out-of-View, Partial Occlusion, Similar Object, Scale Variation and Viewpoint Change. Each sequence carries several of them. 
\begin{table*}[htbp]
\centering
\caption{Attribute based evaluation results. AUC(\%) of DCF-ASN and state-of-the-art trackers on different attributes in the $\textbf{UAV123}$ dataset. The \textcolor[rgb]{ 1,  0,  0}{ﬁrst}, \textcolor[rgb]{ 0,  .69,  .941}{second} and \textcolor[rgb]{ 0,  .69,  .314}{third} highest values are highlighted in color. }
\resizebox{1\linewidth}{!}{
	\begin{tabular}{lccccccccccc}
		\hline
		& \multicolumn{1}{l}{DCF-ASN} &  \multicolumn{1}{l}{DiMP}&\multicolumn{1}{l}{ATOM} & \multicolumn{1}{l}{SiamRPN++} & \multicolumn{1}{l}{SiamMask} & \multicolumn{1}{l}{TADT} & \multicolumn{1}{l}{ARCF} & \multicolumn{1}{l}{VITAL} & \multicolumn{1}{l}{ECO} & \multicolumn{1}{l}{MCCT} & \multicolumn{1}{l}{STRCF} \\
		\hline
		Scale Variation & \textcolor[rgb]{ 1,  0,  0}{64.2} &\textcolor[rgb]{ 0,  .69,  .941}{62.5}& {61.1} & \textcolor[rgb]{ 0,  .69,  .314}{62.3} & 58.2  & 50.6  & 44.2  & 50.6  & 49.4  & 47.2  & 44.6 \\
		
		Aspect Ratio Change & \textcolor[rgb]{ 1,  0,  0}{62.2} &\textcolor[rgb]{ 0,  .69,  .941}{62.0}& {59.9} & \textcolor[rgb]{ 0,  .69,  .314}{61.4} & 56.3  & 47.4  & 41.6  & 47.4  & 45.4  & 44.5  & 39.6 \\
		
		Low Resolution & \textcolor[rgb]{ 1,  0,  0}{51.6} &\textcolor[rgb]{ 0,  .69,  .941}{48.7}& \textcolor[rgb]{ 0,  .69,  .314}{46.7} & {45.4} & 42.4  & 37.2  & 33.9  & 37.2  & 37.6  & 35.6  & 34.1 \\
		
		Fast Motion & \textcolor[rgb]{ 1,  0,  0}{61.4} & \textcolor[rgb]{ 0,  .69,  .941}{61.2}&\textcolor[rgb]{ 0,  .69,  .314}{59.9} & {58.1} & 54.0    & 41.9  & 34.0    & 41.9  & 43.1  & 40.3  & 34.2 \\
		Full Occlusion & \textcolor[rgb]{ 1,  0,  0}{45.0} &\textcolor[rgb]{ 0,  .69,  .941}{44.8}& {40.2} & \textcolor[rgb]{ 0,  .69,  .314}{42.5} & 34.0    & 29.8  & 23.6  & 29.8  & 29.8  & 28.6  & 23.8 \\
		
		Partial Occlusion & \textcolor[rgb]{ 1,  0,  0}{59.7} &\textcolor[rgb]{ 0,  .69,  .941}{58.1}& \textcolor[rgb]{ 0,  .69,  .314}{57.0} & {56.3} & 50.8  & 47.0    & 39.5  & 47.0    & 44.7  & 44.3  & 39.7 \\
		
		Out-of-view & \textcolor[rgb]{ 0,  .69,  .941}{60.8} &\textcolor[rgb]{ 0,  .69,  .314}{59.6}& {58.6} & \textcolor[rgb]{ 1,  0,  0}{60.9} & 56.7  & 46.6  & 39.6  & 46.6  & 43.8  & 43.1  & 38.4 \\
		
		Background Clutter & \textcolor[rgb]{ 1,  0,  0}{48.6} &\textcolor[rgb]{ 0,  .69,  .941}{48.0}&  \textcolor[rgb]{ 0,  .69,  .314}{47.2} &{44.8} & 36.9  & 41.1  & 33.8  & 41.1  & 36.7  & 40.3  & 31.6 \\
		
		Illumination Variation & \textcolor[rgb]{ 0,  .69,  .314}{61.0} &\textcolor[rgb]{ 0,  .69,  .941}{63.0}& \textcolor[rgb]{ 1,  0,  0}{63.1} & {60.7} & 53.7  & 49.6  & 39.9  & 49.6  & 45.9  & 47.8  & 36.7 \\
		
		Viewpoint Change & \textcolor[rgb]{ 0,  .69,  .941}{65.8} & \textcolor[rgb]{ 0,  .69,  .314}{64.8}& \textcolor[rgb]{ 0,  .69,  .314}{64.8} & \textcolor[rgb]{ 1,  0,  0}{68.2} & 62.6  & 50.4  & 41.8  & 50.4  & 48.6  & 46.2  & 40.8 \\
		
		Camera Motion & \textcolor[rgb]{ 1,  0,  0}{66.3} &\textcolor[rgb]{ 0,  .69,  .314}{65.6}& {65.2} & \textcolor[rgb]{ 0,  .69,  .941}{65.8} & 60.6  & 53.6  & 45.4  & 53.6  & 49.9  & 50.1  & 46.5 \\
		
		Similar Object & \textcolor[rgb]{ 1,  0,  0}{63.1} &\textcolor[rgb]{ 0,  .69,  .941}{62.4}& \textcolor[rgb]{ 0,  .69,  .314}{61.0} & {59.0} & 53.9  & 49.7  & 46.0    & 49.7  & 47.7  & 47.3  & 44.4 \\
		\hline
	\end{tabular}%
}
\label{tab:uav123s}%
\end{table*}%

\begin{table*}[htbp]
\centering
\caption{Attribute based evaluation results. Pr(\%) of DCF-ASN and state-of-the-art trackers on different attributes in the  $\textbf{UAV123}$ dataset. The \textcolor[rgb]{ 1,  0,  0}{ﬁrst}, \textcolor[rgb]{ 0,  .69,  .941}{second} and \textcolor[rgb]{ 0,  .69,  .314}{third} highest values are highlighted in color.}
\resizebox{1\linewidth}{!}{
	\begin{tabular}{lccccccccccc}
		\hline
		& \multicolumn{1}{l}{DCF-ASN} &  \multicolumn{1}{l}{DiMP}&\multicolumn{1}{l}{ATOM} & \multicolumn{1}{l}{SiamRPN++} & \multicolumn{1}{l}{SiamMask} & \multicolumn{1}{l}{TADT} & \multicolumn{1}{l}{ARCF} & \multicolumn{1}{l}{VITAL} & \multicolumn{1}{l}{ECO} & \multicolumn{1}{l}{MCCT} & \multicolumn{1}{l}{STRCF} \\
		\hline
		Scale Variation & \textcolor[rgb]{ 1,  0,  0}{84.2} &\textcolor[rgb]{ 0,  .69,  .941}{83.0}& \textcolor[rgb]{ 0,  .69,  .314}{82.8} & {82.0} & 76.9  & 69.2  & 64.0    & 71.3  & 71.3  & 69.8  & 64.0 \\
		
		Aspect Ratio Change &\textcolor[rgb]{ 0,  .69,  .941}{82.8} & \textcolor[rgb]{ 1,  0,  0}{83.3}& \textcolor[rgb]{ 0,  .69,  .314}{82.2} & {81.8} & 75.8  & 66.4  & 61.3  & 68.8  & 67.9  & 67.9  & 58.5 \\
		
		Low Resolution & \textcolor[rgb]{ 1,  0,  0}{75.9} &\textcolor[rgb]{ 0,  .69,  .941}{73.3}& \textcolor[rgb]{ 0,  .69,  .314}{72.6} & {69.0} & 63.8  & 67.7  & 57.0    & 63.5  & 64.5  & 63.3  & 59.7 \\
		
		Fast Motion & \textcolor[rgb]{ 1,  0,  0}{85.0} &\textcolor[rgb]{ 0,  .69,  .941}{83.2}&\textcolor[rgb]{ 0,  .69,  .314} {82.5} & {77.4} & 73.3  & 61.0    & 50.5  & 63.6  & 67.2  & 63.6  & 56.2 \\
		
		Full Occlusion &\textcolor[rgb]{ 0,  .69,  .941}{67.1} & \textcolor[rgb]{ 1,  0,  0}{70.3}& \textcolor[rgb]{ 0,  .69,  .314}{66.7} & {66.1} & 54.7  & 60.5  & 45.3  & 55.9  & 56.1  & 54.8  & 45.3 \\
		
		Partial Occlusion & \textcolor[rgb]{ 1,  0,  0}{81.0} &\textcolor[rgb]{ 0,  .69,  .941}{80.2}& \textcolor[rgb]{ 0,  .69,  .941}{80.2} & \textcolor[rgb]{ 0,  .69,  .314}{77.1} & 70.2  & 69.0    & 57.6  & 68.9  & 66.9  & 67.3  & 58.2 \\
		
		Out-of-view &\textcolor[rgb]{ 0,  .69,  .941}{80.6} & \textcolor[rgb]{ 0,  .69,  .314}{79.9}&{79.4} &  \textcolor[rgb]{ 1,  0,  0}{81.6} & 76.2  & 60.9  & 54.0    & 66.3  & 62.0    & 62.2  & 53.3 \\
		
		Background Clutter & 65.1  & \textcolor[rgb]{ 1,  0,  0}{71.1}&\textcolor[rgb]{ 0,  .69,  .941}{70.9} & \textcolor[rgb]{ 0,  .69,  .314}{65.5} & 55.7  & {68.3} & 56.0    & 63.9  & 59.9  & 62.5  & 50.2 \\
		
		Illumination Variation &  \textcolor[rgb]{ 0,  .69,  .314}{83.1} &\textcolor[rgb]{ 0,  .69,  .941}{85.6}& \textcolor[rgb]{ 1,  0,  0}{87.2} &{81.5} & 73.6  & 66.9  & 60.7  & 73.1  & 71.6  & 73.3  & 55.5 \\
		Viewpoint Change & \textcolor[rgb]{ 0,  .69,  .314}{84.7} &83.7& \textcolor[rgb]{ 0,  .69,  .941}{84.9} & \textcolor[rgb]{ 1,  0,  0}{87.6} & 81.5  & 65.2  & 60.8  & 69.9  & 71.2  & 67.7  & 59.0 \\
		
		Camera Motion & \textcolor[rgb]{ 1,  0,  0}{88.1} &\textcolor[rgb]{ 0,  .69,  .314}{86.7}& \textcolor[rgb]{ 0,  .69,  .941}{86.9} & {86.3} & 79.9  & 71.4  & 64.2  & 74.3  & 70.7  & 71.6  & 65.3 \\
		
		Similar Object &\textcolor[rgb]{ 0,  .69,  .314 }{81.1} &\textcolor[rgb]{ 0,  .69,  .941}{83.7}& \textcolor[rgb]{ 1,  0,  0}{85.0} & {80.0} & 71.6  & 72.2  & 68.0    & 70.5  & 71.3  & 71.8  & 63.1 \\
		\hline
	\end{tabular}%
}
\label{tab:uav123p}%
\end{table*}%

The success and precision plots on the UAV123 dataset are shown in Figure \ref{fig:fig7}, while the further attribute based evaluation results are presented in Table \ref{tab:uav123s} and \ref{tab:uav123p}. Overall, DCF-ASN achieves the highest score of 0.868 and 0.654 on both Pr and AUC among approaches compared. The experimental results indicate that the proposed method are able to estimate the target state in complex aerial scenarios with balanced accuracy and speed.

\paragraph{\textbf{Evaluation on UAVDT Dataset}}
UAVDT is a recent created aerial video dataset that includes three fundamental tasks, i.e., object detection, single object tracking and multiple object tracking. Here, we utilize the test set  of the single object tracking part (50 video sequences). It mainly contains various common scenes of urban district such as squares, arterial streets, toll station, highways, crossings and T-junctions. The video sequences in this dataset are annotated with 9 attributes, that is: Background Clutter, Camera Rotation, Object Rotation, Small Object, Illumination Variation, Object Blur, Scale Variation, Large Occlusion and Long-term Tracking. Notably, about 74\% of video sequences are annotated with at least 4 attributes.
\begin{table*}[htbp]

\centering
\caption{Attribute based evaluation results.  AUC(\%) of DCF-ASN and state-of-the-art trackers on different attributes in the  $\textbf{UAVDT}$ dataset.The \textcolor[rgb]{ 1,  0,  0}{ﬁrst}, \textcolor[rgb]{ 0,  .69,  .941}{second} and \textcolor[rgb]{ 0,  .69,  .314}{third} highest values are highlighted in color.}
\resizebox{1\linewidth}{!}{
	\begin{tabular}{lcccccccccccccc}
		\hline
		& \multicolumn{1}{l}{DCF-ASN} &  \multicolumn{1}{l}{DiMP}&\multicolumn{1}{l}{ATOM} & \multicolumn{1}{l}{SiamRPN++} & \multicolumn{1}{l}{SiamMask} & \multicolumn{1}{l}{TADT} & \multicolumn{1}{l}{ARCF} & \multicolumn{1}{l}{VITAL} & \multicolumn{1}{l}{ECO} & \multicolumn{1}{l}{MCCT} & \multicolumn{1}{l}{STRCF} \\
		\hline
		Camera Motion & \textcolor[rgb]{ 1,  0,  0}{61.5} &57.8& \textcolor[rgb]{ 0,  .69,  .941}{58.9} & 56.9  & \textcolor[rgb]{ 0,  .69,  .314}{57.9} & 41.7  & 44.9  & 41.9  & 41.9  & 40.9  & 41.3 \\
		Object Motion & \textcolor[rgb]{ 1,  0,  0}{60.8} &56.0& 56.7  & \textcolor[rgb]{ 0,  .69,  .941}{59.4} & \textcolor[rgb]{ 0,  .69,  .314}{57.0} & 38.2  & 41.7  & 40.9  & 38.2  & 39.2  & 38.6 \\
		Small Object & \textcolor[rgb]{ 0,  .69,  .314}{57.2} &56.4& \textcolor[rgb]{ 0,  .69,  .314}{57.2} & \textcolor[rgb]{ 0,  .69,  .941}{59.2} & \textcolor[rgb]{ 1,  0,  0}{60.9} & 44.5  & 49.2  & 44.8  & 44.0    & 45.0    & 45.0 \\
		Illumination Variations & \textcolor[rgb]{ 0,  .69,  .314}{61.3} &58.8& 58.8  & \textcolor[rgb]{ 1,  0,  0}{66.4} & \textcolor[rgb]{ 0,  .69,  .941}{63.4} & 44.4  & 47.8  & 48.7  & 44.6  & 47.4  & 47.8 \\
		Object Blur & \textcolor[rgb]{ 0,  .69,  .314}{59.3} &54.9& 55.1  & \textcolor[rgb]{ 1,  0,  0}{65.8} & \textcolor[rgb]{ 0,  .69,  .941}{61.8} & 43.2  & 46.8  & 46.1  & 43.4  & 46.4  & 45.8 \\
		Scale Variations & \textcolor[rgb]{ 1,  0,  0}{63.4} &58.3& \textcolor[rgb]{ 0,  .69,  .941}{60.3} & \textcolor[rgb]{ 0,  .69,  .314}{60.1} & 58.5  & 43.2  & 43.7  & 43.6  & 40.7  & 41.0    & 41.4 \\
		Long-term Tracking & \textcolor[rgb]{ 1,  0,  0}{72.4} & \textcolor[rgb]{ 0,  .69,  .314}{70.5}&{70.4} & 65.1  & \textcolor[rgb]{ 0,  .69,  .941}{70.7} & 52.9  & 57.9  & 55.3  & 55.3  & 60.3  & 60.8 \\
		Background Clutter & \textcolor[rgb]{ 1,  0,  0}{56.1} &52.5& \textcolor[rgb]{ 0,  .69,  .314}{52.9} & \textcolor[rgb]{ 0,  .69,  .941}{54.7} & 52.3  & 40.1  & 41.4  & 39.0    & 36.1  & 38.3  & 37.5 \\
		Large Occlusion & \textcolor[rgb]{ 0,  .69,  .941}{55.2} &\textcolor[rgb]{ 1,  0,  0}{55.6}& \textcolor[rgb]{ 1,  0,  0}{55.6} & \textcolor[rgb]{ 0,  .69,  .314}{49.8} & 46.0    & 39.4  & 38.7  & 36.7  & 34.2  & 33.1  & 32.8 \\
		\hline\\
	\end{tabular}%
}
\label{tab:uavdts}%
\end{table*}%

Figure \ref{fig:fig8} presents the comparison results of the proposed DCF-ASN and aforementioned preeminent algorithms on the UAVDT datasets. Obviously, our DCF-ASN method gains the first in terms of precision and AUC with the score 83.7\% and 61.4\%, respectively. As reported in Table \ref{tab:uavdts} and \ref{tab:uavdtp}, the proposed approach always ranks top-3 on all challenging attributes, which proves its robustness and accurancy.
\begin{table*}[htbp]
\centering
\caption{Attribute based evaluation results.  Pr(\%) of DCF-ASN and state-of-the-art trackers on different attributes in the  $\textbf{UAVDT}$ dataset. The \textcolor[rgb]{ 1,  0,  0}{ﬁrst}, \textcolor[rgb]{ 0,  .69,  .941}{second} and \textcolor[rgb]{ 0,  .69,  .314}{third} highest values are highlighted in color.}
\resizebox{1\linewidth}{!}{
	\begin{tabular}{lcccccccccccccc}
		\hline
		& \multicolumn{1}{l}{DCF-ASN} &  \multicolumn{1}{l}{DiMP}&\multicolumn{1}{l}{ATOM} & \multicolumn{1}{l}{SiamRPN++} & \multicolumn{1}{l}{SiamMask} & \multicolumn{1}{l}{TADT} & \multicolumn{1}{l}{ARCF} & \multicolumn{1}{l}{VITAL} & \multicolumn{1}{l}{ECO} & \multicolumn{1}{l}{MCCT} & \multicolumn{1}{l}{STRCF} \\
		\hline
		Camera Motion & \textcolor[rgb]{ 1,  0,  0}{84.0} &79.0& \textcolor[rgb]{ 0,  .69,  .941}{80.0} &  75.9  & \textcolor[rgb]{ 0,  .69,  .314}{76.7} & 64.7  & 71.5  & 70.4  & 67.3  & 65.2  & 64.5 \\
		Object Motion & \textcolor[rgb]{ 1,  0,  0}{82.6} &77.3& 77.5  & \textcolor[rgb]{ 0,  .69,  .941}{80.4} & \textcolor[rgb]{ 0,  .69,  .314}{77.8} & 62.5  & 66.7  & 67.7  & 62.7  & 60.8  & 60.2 \\
		Small Object & \textcolor[rgb]{ 0,  .69,  .941}{86.5} &81.5& 83.3  & 83.5  & \textcolor[rgb]{ 1,  0,  0}{86.7} & 81.6  & \textcolor[rgb]{ 0,  .69,  .314}{84.8} & 81.4  & 80.1  & 80.4  & 78.5 \\
		Illumination Variations & \textcolor[rgb]{ 0,  .69,  .314}{84.5} &81.7& 82.2  & \textcolor[rgb]{ 1,  0,  0}{89.7} & \textcolor[rgb]{ 0,  .69,  .941}{86.4} & 76.1  & 79.6  & 82.4  & 77.1  & 77.9  & 77.5 \\
		Object Blur & \textcolor[rgb]{ 0,  .69,  .314}{82.7} &75.9& 77.6  & \textcolor[rgb]{ 1,  0,  0}{89.4} & \textcolor[rgb]{ 0,  .69,  .941}{86.0} & 74.4  & 75.9  & 75.9  & 75.3  & 75.9  & 74.5 \\
		Scale Variations & \textcolor[rgb]{ 1,  0,  0}{82.4} &77.7& \textcolor[rgb]{ 0,  .69,  .314}{78.4} &  \textcolor[rgb]{ 0,  .69,  .941}{80.1} & 77.3  & 61.0    & 64.2  & 65.3  & 60.5  & 59.0    & 58.0 \\
		Long-term Tracking & {97.0} & \textcolor[rgb]{ 1,  0,  0}{99.9}&\textcolor[rgb]{ 0,  .69,  .314} {97.2} & 84.9  & 93.8  & 77.4  & 88.3  & 90.7  & 89.1  &\textcolor[rgb]{ 0,  .69,  .941}{98.6} & 93.9 \\
		Background Clutter & \textcolor[rgb]{ 1,  0,  0}{76.6} &\textcolor[rgb]{ 0,  .69,  .314}{73.5}& {72.2}  & \textcolor[rgb]{ 0,  .69,  .941}{74.9} & 71.6  & 66.0    & 66.7  & 67.5  & 59.9  & 58.9  & 59.1 \\
		Large Occlusion & \textcolor[rgb]{ 0,  .69,  .314}{70.5} &\textcolor[rgb]{ 1,  0,  0}{72.3}& \textcolor[rgb]{ 0,  .69,  .941}{72.2} & {66.6} &  60.2  & 52.8  & 54.8  & 52.7  & 48.5  & 41.2  & 42.9 \\
		\hline& \\
	\end{tabular}%
}
\label{tab:uavdtp}%
\end{table*}%

As illustrated in Tables 4-7, DCF-ASN performs well when the camera is fast moving or the object suffers from motion blur, benefitting from the proposed "coarse-to-fine" strategy. The DCF sub-module enables dense sampling around the peak of response map, while less proposals are generated on the positions with a low response value. Thus, the ASN sub-module can make multiple confirmations in the region where the target is most likely to appear. Even if the target scale changes drastically, this coarse-to-fine strategy could helps ensure the robust tracking performance. Both the conventional DCF based algorithms (e.g. ECO) and the popular CNN-based trackers (e.g. DiMP) cannot perform well alone when such challenges exist.

\begin{figure}[!htb]
\centering
\subfigure[AUC \textit{vs} Model Size]{\includegraphics[width=0.48\textwidth]{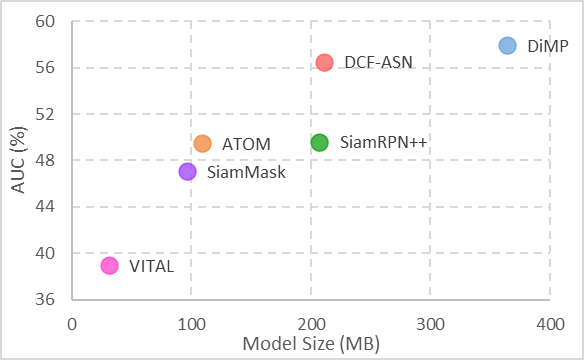}} 
\subfigure[Pr \textit{vs} Model Size]{\includegraphics[width=0.48\textwidth]{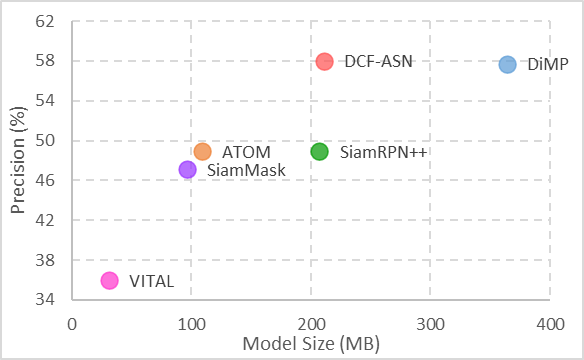}} \\
\subfigure[AUC \textit{vs} Model Size]{\includegraphics[width=0.48\textwidth]{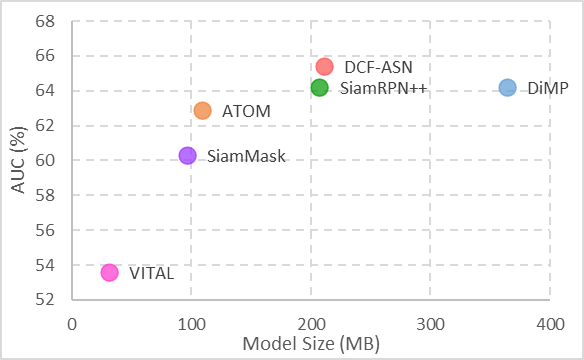}} 
\subfigure[Pr \textit{vs} Model Size]{\includegraphics[width=0.48\textwidth]{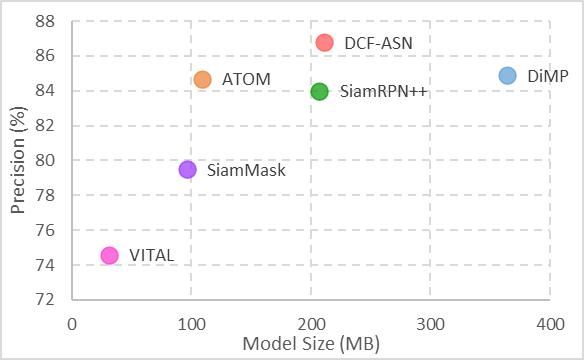}} \\
\caption{Performance vs. number of parameters. Results are evaluated on LaSOT (Figure 9(a) and Figure 9(b)) and UAV123 ( Figure 9(c) and Figure 9(d)) }
\label{fig:fig9}
\vspace{0.2in}
\end{figure}
\begin{figure*}[!htb]
\centering
\subfigure[Sequence \textit{Diving}]{\includegraphics[width=0.95\textwidth]{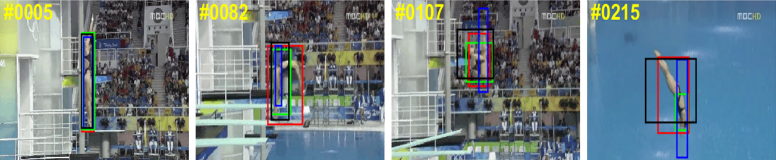}} \\
\subfigure[Sequence \textit{Car16}]{\includegraphics[width=0.95\textwidth]{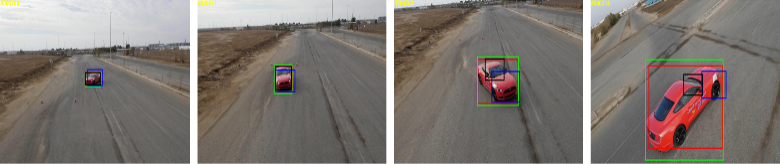}} \\
\includegraphics[width=0.95\linewidth]{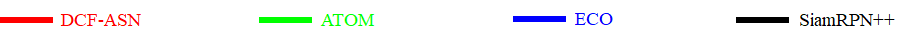}
\caption{Qualitative demonstration of proposed DCF-ASN and representative state-of-the-art (SOTA) trackers on sequences \textit{Diving} and \textit{Car16} obtained from datasets OTB100 and UAV123.}
\label{fig:fig10}
\vspace{0.2in}
\end{figure*}
\subsection{Analyses on Performance and Model Size}
In order to test the efficiency of model parameters usage, comparisons regarding the relationships between performance and model size are done on the large-scale general tracking dataset LaSOT and typical aerial tracking dataset UAV123. Results are shown in Figure \ref{fig:fig9}. Our DCF-ASN achieves competing AUC and Pr on two datasets, having a better tradeoff between performance and model size. Although the number of parameters employed by VITAL is much smaller than that used by others, it can only run at about 3 FPS.  The compression of model size is achieved at the cost of reducing speed, which limits the application of VITAL. Compared with DiMP, our tracker performs better in most cases with less parameters. This indicates that DCF-ASN utilises parameters with high-efficiency, and therefore is more memory efficient.
\subsection{Qualitative evaluation}
To evaluate the performance of the proposed DCF-ASN more comprehensively, we have selected the representative DCF tracker (ECO), the siamense algorithms (SiamRPN++), and the competitive SOTA method (ATOM) to  conduct a qualitative analysis on the sequences in different scenarios. 

As shown in Figure \ref{fig:fig10}, ATOM, though including components for target estimation and classiﬁcation, cannot generate an accurate bounding box. ECO, constantly updated online, suffers from serious model drift and thus loses the precise location and scale of the target. SiamRPN++, an ofﬂine-trained generative tracker, lacks of ability of adopting to the target appearance variation and hence, fails under this circumstance. The proposed DCF-ASN, combining the unique strengths of traditional DCF and siamense tracker, thereby achieving the best tracking accuracy overall.

\section{Conclusion}
This paper has presented a novel coarse-to-fine visual tracking framework, which consists of an online-updating DCF module and an offline-training attentional siamese network (ASN). The proposed ASN is an asymmetric network that can learn to capture the coefficient of different channels from the template and weigh the features of the search region. By integrating a standard DCF with the ASN efficiently, the proposed approach achieves favorable tracking accuracy while holding its real-time ability. Comprehensive experiments on five large-scale datasets have systematically demonstrated the superior performance of this tracking framework. In the future, tensor decomposition techniques may be introduced to reduce the computational burden and speed up the tracker further. Meanwhile, we will continue to improve the precision of bounding box regression by exploiting the advanced anchor-free technique in tracking.

\bibliographystyle{unsrtnat}
\bibliography{references}  






\end{document}